\documentclass[conference]{IEEEtran}
\usepackage{cite}
\ifCLASSINFOpdf
   \usepackage[pdftex]{graphicx}
\else
\fi
\usepackage{amsmath}
\usepackage{array}
\newcommand\Mark[1]{\textsuperscript#1}
\usepackage{stfloats}
\usepackage{url}
\newcommand{\x}{0.35cm}
\newcolumntype{P}[1]{>{\centering\arraybackslash}p{#1}}
\newcolumntype{M}[1]{>{\centering\arraybackslash}m{#1}}
\renewcommand\IEEEkeywordsname{Keywords}

\begin{document}
\title{Applying Data Augmentation to Handwritten Arabic Numeral Recognition Using Deep Learning Neural Networks}
\author{\IEEEauthorblockN{Akm Ashiquzzaman\Mark{1},
		Abdul Kawsar Tushar\Mark{1}, and
		Md Ashiqur Rahman\Mark{2}
	}
	\IEEEauthorblockA{\Mark{1}Computer Science and Engineering Department, University of Asia Pacific, Dhaka, Bangladesh}
	\IEEEauthorblockA{\Mark{2}Computer Science and Engineering Department, Bangladesh University of Engineering and Techology, Dhaka, Bangladesh\\\{zamanashiq3, tushar.kawsar, ashiqbuet14\}@gmail.com}
}
\maketitle

\begin{abstract}
Handwritten character recognition has been the center of research and a benchmark problem in the sector of pattern recognition and artificial intelligence, and it continues to be a challenging research topic. A convolutional neural network model for recognizing handwritten numerals in Arabic language is proposed in this paper, where the dataset is subject to various augmentation in order to add robustness needed for deep learning approach. The proposed method is empowered by the presence of dropout regularization to do away with the problem of data overfitting. Moreover, suitable change is introduced in activation function to overcome the problem of vanishing gradient. With these modifications, the proposed system achieves an accuracy of 99.4\% which performs better than every previous work on the dataset.
\end{abstract}

\IEEEpeerreviewmaketitle

\section{Introduction}


Automatic character recognition, also known as optical character recognition (OCR), has very high commercial and pedagogical importance and has been receiving a profound interest from researchers over the past few years. And this process is being  applied  to different languages all over the world.

Arabic is the fifth most spoken language in the world. Along with 290 million native speakers, it is spoken by 422 million speakers in 22 countries \cite{unesco}. Arabic is the major source of vocabulary for languages Turkish, Uighur, Urdu, Kazakh, Kurdish, Uzbek, Kyrgyz, and Persian. \cite{boucenna2006origin}.

\begin{figure}[!t]
\centering{\includegraphics[width=80mm]{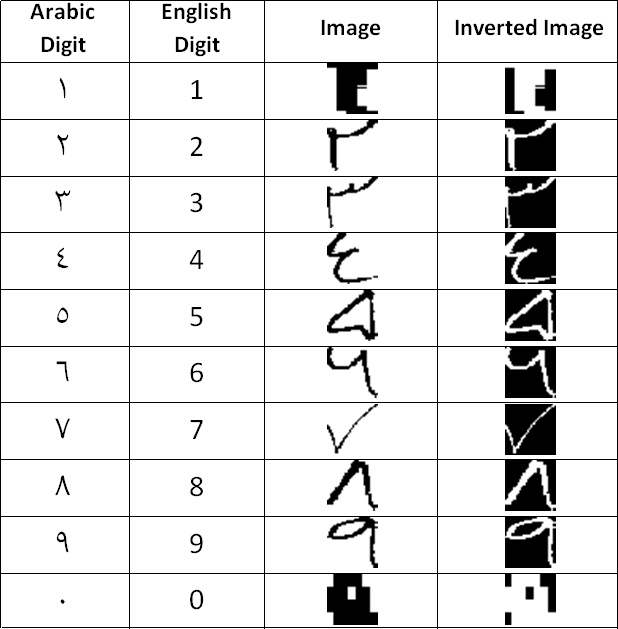}}
	\caption{Handwritten Arabic digits and corresponding inverted images \cite{akshiquzzaman}}
	\label{fig:table}
\end{figure}

The past works of OCR of hand written alphabets and numerals were concentrated on Latin languages. Gradually other languages are also coming to the fore. But Arabic still has many unexplored fields of study. Maximum work done on this language was to detect printed characters \cite{amin1998off}; however, detection of hand written alphabets are more challenging than printed characters. Some work has been done on Arabic handwritten digit recognition via patterns leanred from other related languages \cite{tushar2017novel}. The interesting point of note regarding Arabic is, though the words as well as characters are written from right to left, the numerals are written from left to right. Fig. \ref{fig:table} depicts the ten Arabic numeral classes.

Das et al. \cite{Nibaran} have devised an method to recognize handwritten Arabic numerals by multilayer perceptron (MLP) with a considerable accuracy. In \cite{ashiquzzaman} a convolutional neural network (CNN) model \cite{krizhevsky2012imagenet} is used with dropout based on the model \cite{Nibaran} that achieves a better accuracy on the same dataset. In this paper, we propose a modification to the same method devised in \cite{ashiquzzaman}. We introduce data augmentation \cite{krizhevsky2012imagenet} to make learning more robust against overfitting. Exponential linear unit (ELU) is introduced instead of Rectified Linear Unit (ReLU) to fix the vanishing gradient problem.

\begin{figure*}[!tb]
	\centerline{\includegraphics[width=\textwidth, height=40mm]{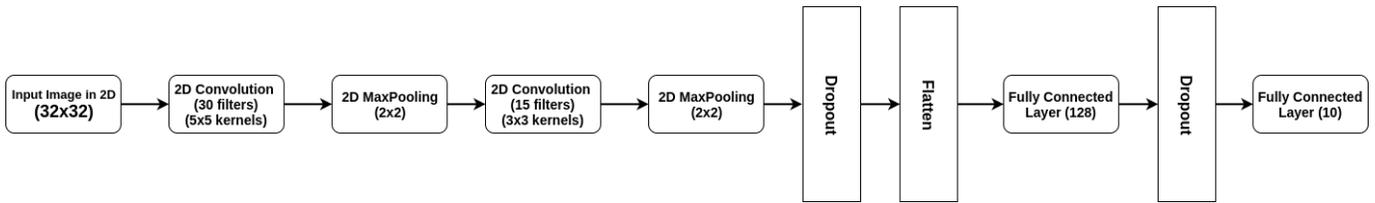}}
	\caption{CNN used in \cite{ashiquzzaman} }
	\label{fig:cnn}
\end{figure*}
\begin{figure*}[b]
	\centerline{\includegraphics[width=\textwidth]{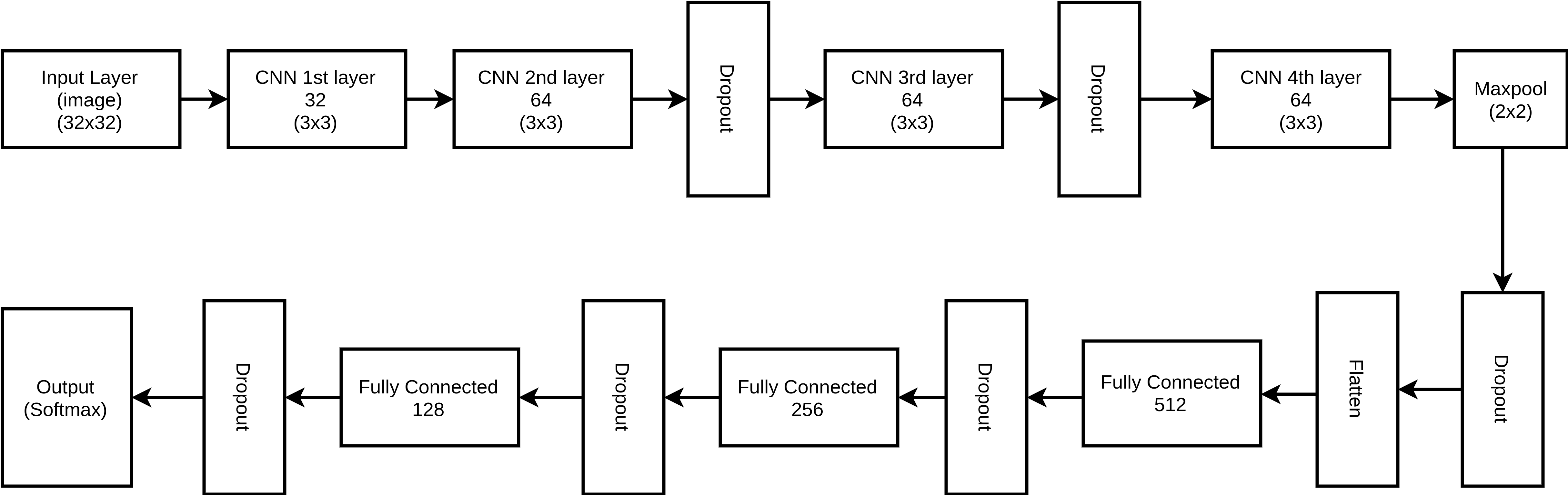}}
	\caption{Proposed CNN }
	\label{fig:cnn1}
\end{figure*}

\section{Background}\label{background}

The method devised in \cite{Nibaran} to recognize Arabic numerals uses MLP as a classifier. It uses a set of 88 features; among them 72 are shadow feature and 16 are octant features. To train this MLP a huge data set of 3000 handwritten sample is used which is obtained from  CMATERDB 3.3.1 \cite{web4}. Each of these pictures is scaled to the size of $32\times32$. For uniform training each pixel is redefined in gray scale.

\begin{figure}[bh]
\centerline{\includegraphics[width=80mm, height=40mm]{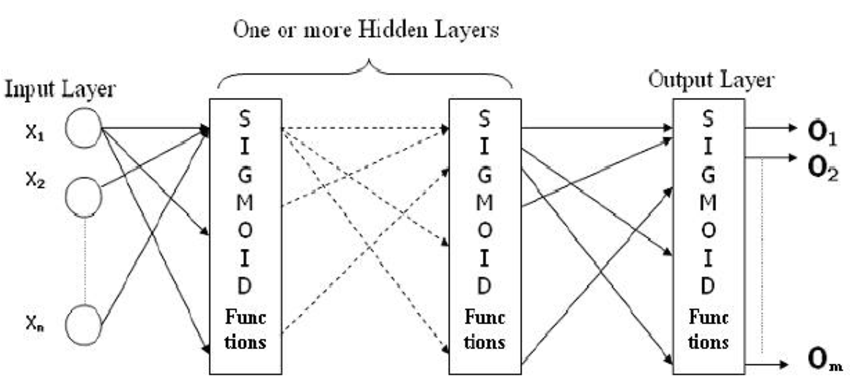}}
	\caption{MLP used by Das et al. \cite{Nibaran}.}
	\label{fig:mlp}
\end{figure}

The MLP used in \cite{Nibaran} has a single hidden layer along with input and output layer. And this single hidden layer is enough to classify the given data set \cite{DBLP:journals/ker/Kubat99a}. The setting of 3 layers for the MLP method is shown in Fig. \ref{fig:mlp}. From the Fig. \ref{fig:mlp} we can see that sigmoid is used as nonlinearity after each layer. The width of the network is determined through trial and error to get a good bias and minimal variance. Supervised learning is undertaken and performed over 2000 samples. Simple back-propagation algorithm is used for training.

The result was evaluated by cross validation, and each time the number of neurons are varied to get a perfect bias-variance combination. By exchanging the number of neurons arbitrarily between training data set and test data set error can be brought very close to zero, but this will generate worse performance when tested on samples outside of sample dataset. This condition is known as ``overfitting" \cite{DBLP:conf/icml/Domingos00} and this is because the model gets too much acquainted to the training data set and cannot generalize over a broader spectrum as a result. The number of neurons in hidden layer is finally fixed to 54 that gives an accuracy of 93.8\% in recognizing Arabic numerals.

Das et al. split the total data set at a ratio 2:1 for training and testing. The images are normalized before feeding into the network. Before feeding data into MLP the images were transformed into a simple one dimensional vector.

\begin{figure}[!b]
	\centerline{\includegraphics[width=100mm]{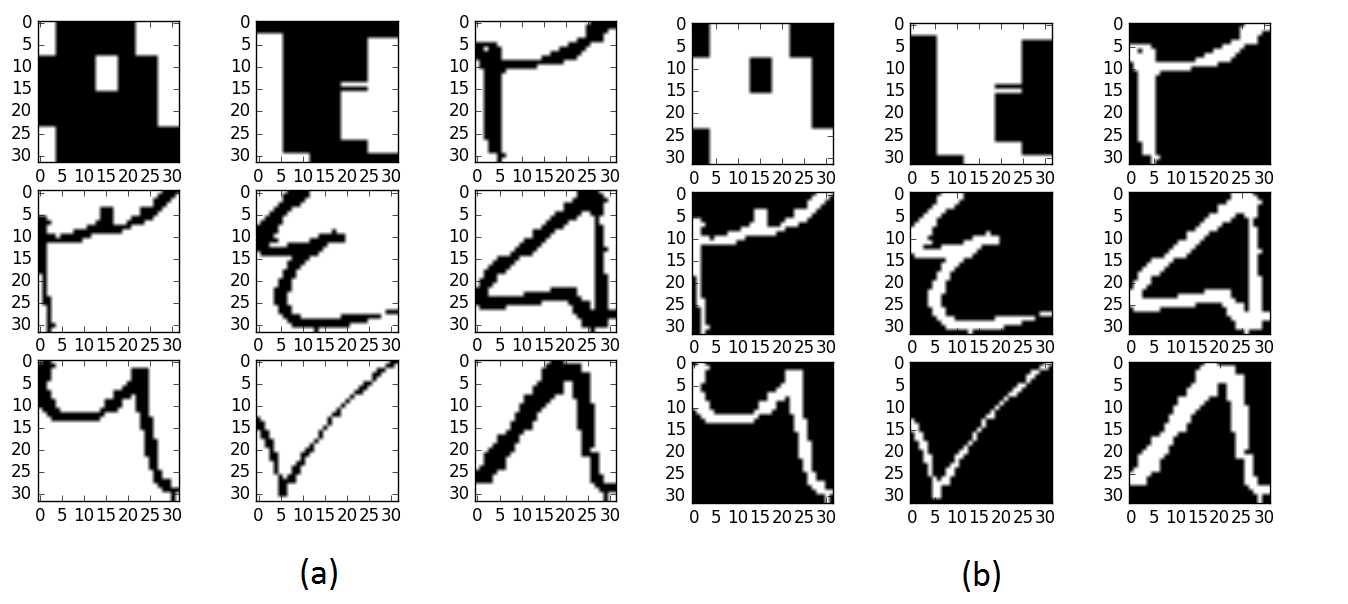}}
	\caption{(a) Original Data. (b) Final Data after processing}
	\label{fig:raw-v-preprocessed}
\end{figure}

The model described in \cite{ashiquzzaman} uses the same dataset as \cite{Nibaran} but adopts the method of CNN for numeral recognition. The images are kept in the original form for CNN; however, color of the images are inverted before feeding. Fig. \ref{fig:raw-v-preprocessed} shows the final condition of images before feeding in CNN model. Fig. \ref{fig:cnn} describes the total model used by them. From the Fig. \ref{fig:cnn} it is seen that, their model consists of two layers of convolution with  kernels of size 5x5 and 3x3, respectively each followed by a 2x2 maxpooling layer. The fully connected network consists of a hidden layer with 128 neurons and one final layer with 10 neurons for each numeral class. They use ReLU \ref{activation_function} as activation function and categorical cross entropy for error calculation. Softmax function is used to get the final result from output layer. Total dataset is split at ratio 2:1 for training and validation purpose. This model gives an accuracy of 97.4\% over validation dataset.


\section{Proposed Method}
\begin{table*}[t]
	\centering
	\caption{Some well known activation functions \cite{clevert2015fast} \cite{nair2010rectified}}
		\renewcommand{\arraystretch}{1.5}
        \begin{small}
		{\begin{tabular}{|>{\centering\arraybackslash}m{0.5in} |>{\centering\arraybackslash}m{1.2in} |>{\centering\arraybackslash}m{2.35in} |>{\centering\arraybackslash}m{2.35in} |}
				\hline
				Name & Plot & Equation & Derivative \\ \hline
                Logistic  & \includegraphics[width=30mm,height=15mm]{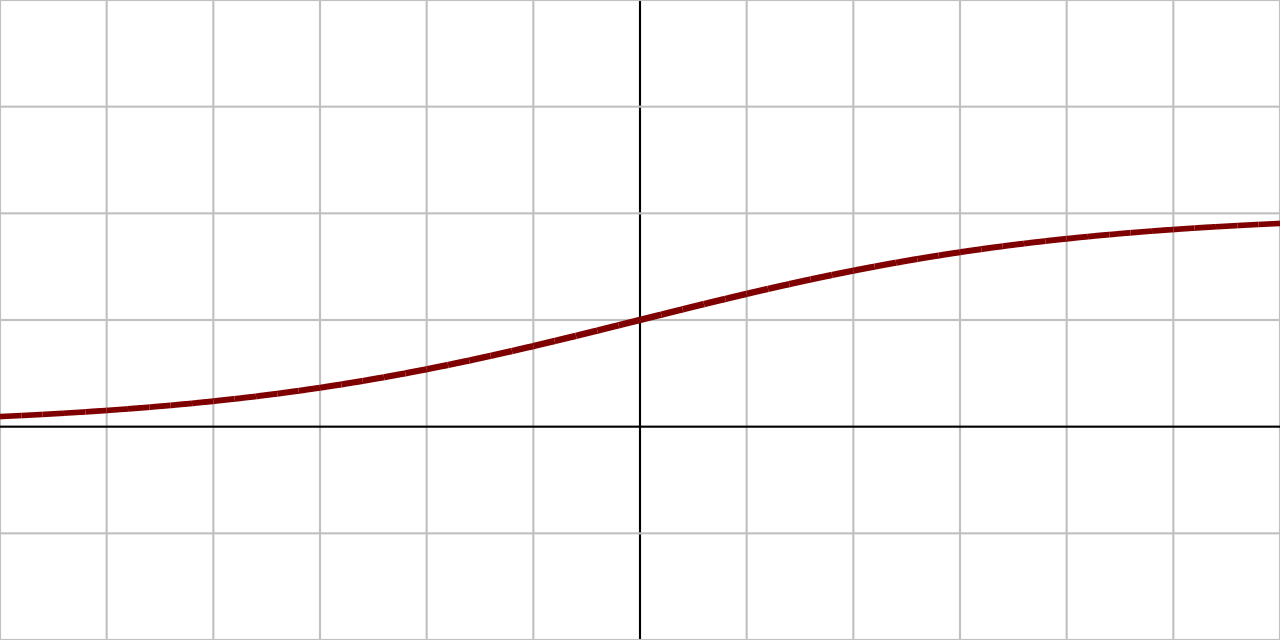} & $ f(x)=\frac{1}{e^{-x}+1}$ & $f'(x)=f(x)\left(f(x)-1\right)$   \\ \hline
                tanh & \includegraphics[width=30mm,height=15mm]{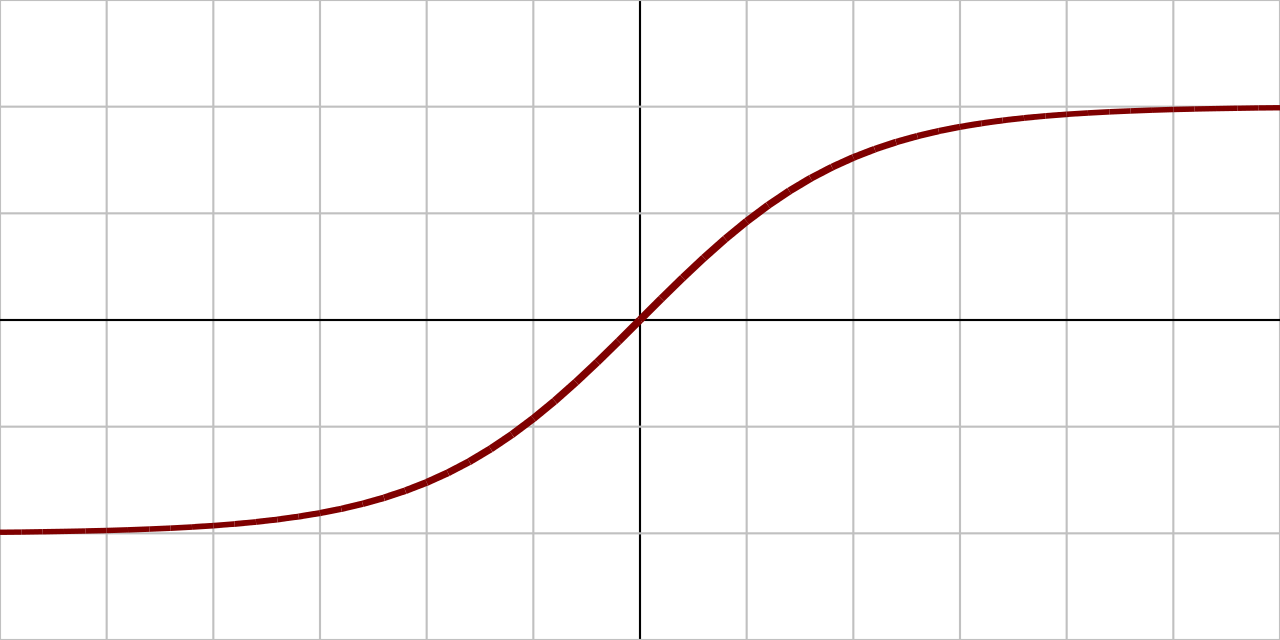} & $ f(x)= \tanh(x) $  & $f'(x)=1-f(x)^2$ \\ \hline
                Arctan & \includegraphics[width=30mm,height=15mm]{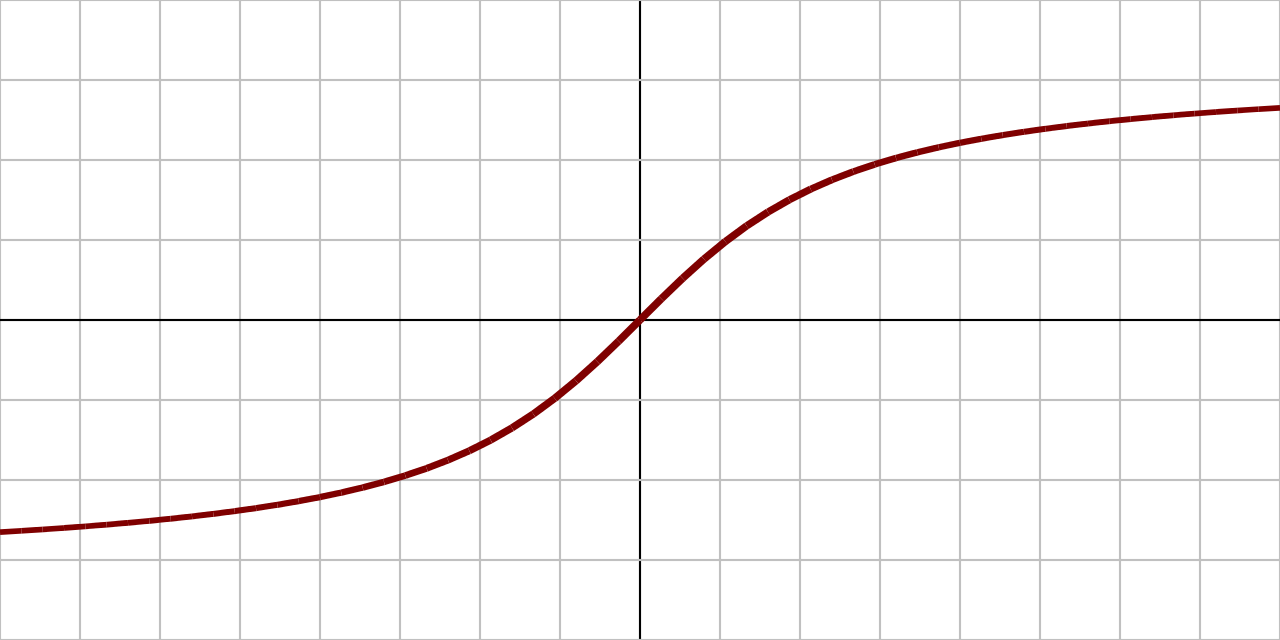} & $f(x)=\tan^{-1}(x) $ & $f'(x)= \frac{1}{1+x^2}$ \\ \hline
                ReLU & \includegraphics[width=30mm,height=15mm]{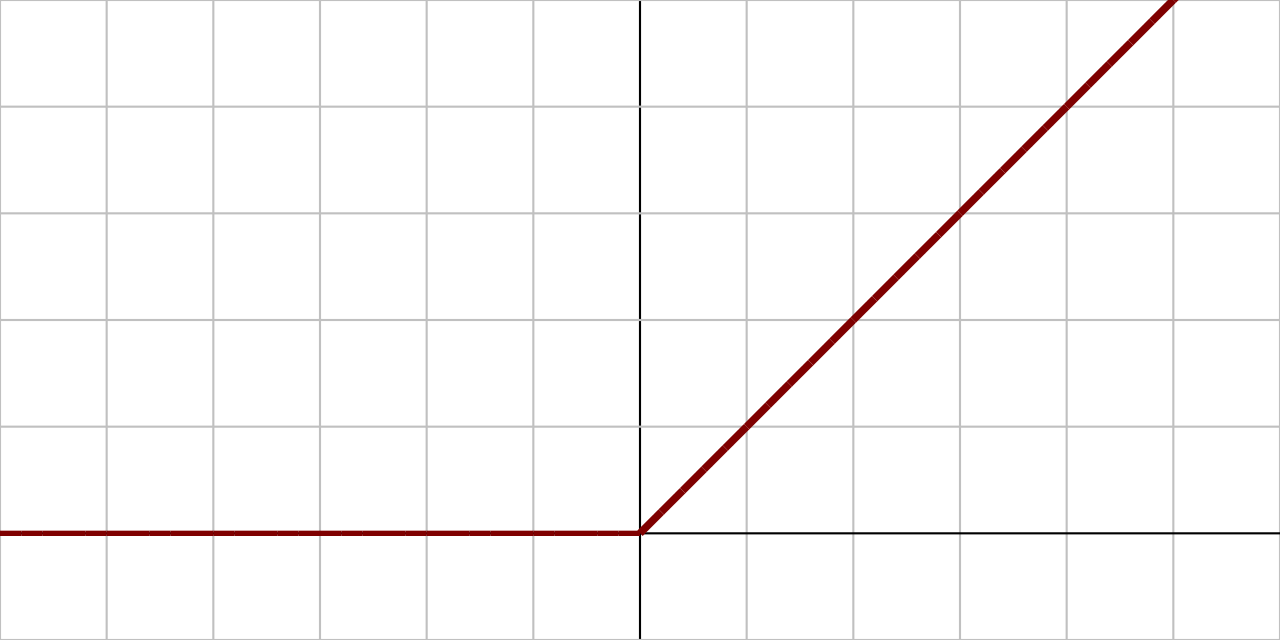} & $f(x) = \left\{
                		\begin{array}{lr}
                		0 & for~ x < 0\\
               			x & for~ x \ge 0
                		\end{array}
                			\right.$	&  $f'(x) = \left\{
                		\begin{array}{lr}
                		0 & for~ x < 0\\
               			1 & for~ x \ge 0
                		\end{array}
                			\right.$ \\ \hline
                 ELU & \includegraphics[width=30mm,height=15mm]{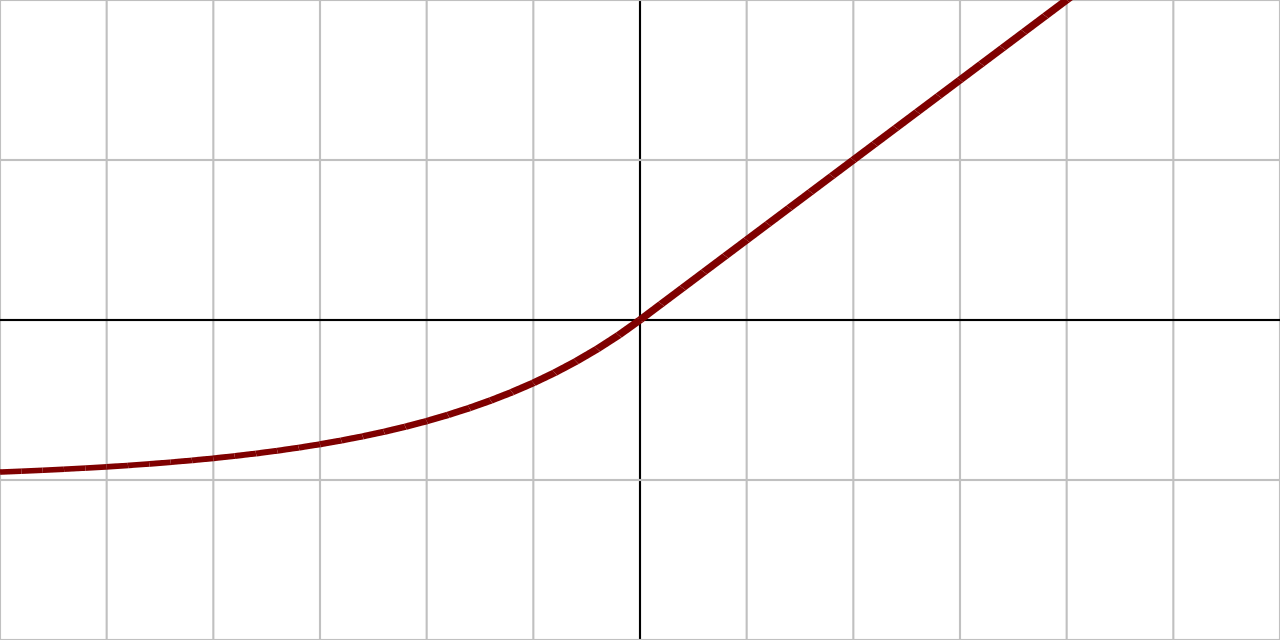}& $f(\alpha,x) = \left\{
                		\begin{array}{lr}
                		\alpha(e^x-1) & for~ x < 0\\
               			x & for~ x \ge 0
                		\end{array}
                			\right.$	&  $f'(\alpha, x) = \left\{
                		\begin{array}{lr}
                		f(\alpha,x)+\alpha & for~ x < 0\\
               			1 & for~ x \ge 0
                		\end{array}
                			\right.$ \\ \hline		
		\end{tabular}}{\label{activation_function}}	
        \end{small}
\end{table*}

In this section we discuss the changes that we bring to the method described in \cite{ashiquzzaman} to improve the accuracy. Two key changes are brought about - we introduce data augmentation to the CMATERDB 3.3.1 dataset, and change the activation function from ReLU to ELU. The pictorial presentation of our model is given in Fig. \ref{fig:cnn1}.

Data augmentation \cite{krizhevsky2012imagenet} transforms our base data. In our case, it simply takes our dataset of images and transforms it by rotation, color variation or by adding noise. It makes our model more robust against overfitting and numerically increases the size of the dataset. We've apply ZCA-whitening as augmentation, images are also rotated in a range of 10 degree. The images are shifted both horizontally and vertically to an extent of 20\% of the original dimensions randomly. The images are zoomed randomly up to 10\%. During this augmentation process the points outside the boundaries are filled according to the nearest point.\\
\begin{eqnarray}
f'(x) = \left\{
                		\begin{array}{lr}
                		0 & for~ x < 0\\
               			1 & for~ x \ge 0
                		\end{array}
                			\right. \label{eq:RELU}\end{eqnarray}\\
\begin{eqnarray}                            
f'(\alpha, x) = \left\{
                		\begin{array}{lr}
                		f(\alpha,x)+\alpha & for~ x < 0\\
               			1 & for~ x \ge 0
                		\end{array}
                			\right. \label{eq:ELU}\end{eqnarray}
The ReLU is sometimes plagued with the gradient vanishing problem. From Equation. \ref{eq:RELU} it is seen that for region $x<0$ the derivation of ReLU is $0$, and due to this in this region the updating of weight vector stops. And this stops the learning process. The ELU function can prevent this condition as its derivative Equation \ref{eq:ELU} does not become zero in any point on the curve. Furthermore, it assures a smooth learning .

In our model, we have four convolution layers, each with ELU as an activation function. The kernel size is determined through trial and error, and the window of $3 \times 3$ gives maximum accuracy. Next to the final layer of convolution, we define a pooling layer doing max-pool with a pool size of $2 \times 2$. After max-pooling the convoluted images are flattened by squashing 2D convoluted data and fed into the fully connected layers. The final output layer contains 10 output nodes representing 10 classes of numerals. Softmax  function is used to calculate final result from the output layer. This function calculates the probability of each class from the ELU value of each output layer neuron \cite{hinton2009replicated}.

Both the convolution and fully connected layers have a dropout \cite{srivastava2014dropout} rate of 25\% to prevent data overfitting problem. In this method, in each epoch 25\% neurons in each layers do not update their weight vectors, and the effect of these neurons are removed from network. If neurons are dropped, they are prevented from co-adapting too much with the training set and reduces overfitting.

\section{Experiment}
The success of any machine learning method largely depends on the size and correctness of dataset used. For deep learning the effect of data set is even more vital. The CMATERDB 3.3.1 Arabic handwritten digit dataset is used \cite{web4}. It contains 3000 separate handwritten numeral image. Each of them is 32x32 pixel RGB image. Similar to the process in \cite{ashiquzzaman}, we invert the images before feeding into CNN. As a result the numerals are in white foreground on the backdrop of black background. For the past works done on OCR, it is observed through activation visualization that edges are a very important feature in character recognition, and black background makes the edge detection easier.

We train the proposed CNN model with CMATERDB Arabic handwritten digit dataset and test against the test data sample of same dataset. Similar to the process implementation described in \cite{ashiquzzaman}, our CNN have increased $4,977,290$ instead of total $75,383$ trainable parameters such as weights and biases. Model is implemented in Keras \cite{chollet2015}. Experimental model was implemented in Python using Theano \cite{2016arXiv160502688short} and Keras libraries. The model is trained for 100 epoch with different kernel sizes. The batch size is 128 for both training and testing. Categorical crossentropy is used as the loss function in this model. Adadelta optimizer \cite{zeiler2012adadelta} is used to optimize the learning process. Among the 3000 images, 2500 are used for training and 500 is used in validation. We have used a computer with CPU intel i5-6200U CPU @ 2,30GHz and @ 4GB ram. Nvidia Geforce GTX 625M dedicated graphics is used for faster computation, i.e CUDA support for accelerated training is adopted.

\section{Result And Discussion}\label{result}
Table \ref{table_database} shows the result of different models applied to the same CMATERDB dataset. Method-1 denotes the method proposed in \cite{Nibaran} and Method-2 denotes the method proposed in \cite{ashiquzzaman}.

\begin{table}[t]
	\centering
	\caption{Performance Comparison of Proposed Methods and Methods Described in \cite{ashiquzzaman} and \cite{Nibaran}}
	\begin{small}
		\renewcommand{\arraystretch}{1.5}
		{\begin{tabular}{|l|c|}
				\hline
				\bfseries Method Name & \bfseries Accuracy \\ \hline
				
				 Method-1 & 93.8 \\ \hline
				 Method-2 (without data augmentation) & 97.4 \\ \hline
                Proposed Method ( with data augmentation)  & 99.4 \\ \hline
		\end{tabular}}{\label{table_database}}
	\end{small}
\end{table}

\begin{figure}[h]
\centerline{\includegraphics[width=75mm, height=80mm]{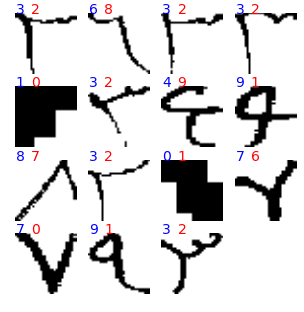}}
	\caption{Some misclassified images during training. Blue denotes actual class and red denotes predicted class.}
	\label{fig:wrong_png}
\end{figure}

Our proposed method of CNN using data augmentation gives better accuracy than the method described in \cite{ashiquzzaman} with no use of data augmentation. The prominent attribute of the deep learning is that deep embedded layers recognize the features and cascade them into the final output prediction to cast classification. This is the attribute that has contributed to deep learning method being used in problems in image recognition and disease prediction \cite{ashiquzzaman2017reduction}, among others. In our proposed model, the CNN layers have been increased in number from the model used by \cite{ashiquzzaman}. We have also added two other fully connected layers in later phase of the proposed model to enhance feature extraction and recognition. The result was significantly improved from the previous methods because of the introduction of data augmentation during training. Data augmentation virtually makes the image transformed into new set of features for the neural network to detect and recognize. It also shifts and zooms the training images, making the image decentralized for the kernels to recognize them from various positions.

Fig. \ref{fig:wrong_png} denotes some of the misclassified images during training. It demonstrates some of misclassified images, which are actually impossible to recognize manually due to morphological decomposition of figure shape. Table \ref{confusion} shows the class-wise classification of the test images, which denotes the final accuracy rate of 99.40\% . The columns denote true classes and the rows denote predicted classes. The cell content denotes the count of predicted classes.

\begin{table}[h]
	\centering
	\caption{Confusion Matrix}
    \label{confusion}
	\begin{small}
		\renewcommand{\arraystretch}{1.5}
		{\begin{tabular}{|M\x|M\x|M\x|M\x|M\x|M\x|M\x|M\x|M\x|M\x|M\x|}
				\hline &
				\bfseries 0 & \bfseries 1&
                \bfseries 2& \bfseries3 &\bfseries 4 &\bfseries 5& \bfseries 6&\bfseries 7& \bfseries 8 &\bfseries 9\\ \hline
			\bfseries0 & 50 & 0 & 0 &	0 & 0 &0 & 0 &0 & 0 &0 \\ \hline
            \bfseries1 & 0 & 50 & 0 &	0 & 0 &0 & 0 &0 & 0 &0 \\ \hline
    \bfseries        2 & 0 & 0 & 49 &	1 & 0 &0 & 0 &0 & 0 &0 \\ \hline
	\bfseries		3 & 0 & 0 & 0 &	50 & 0 &0 & 0 &0 & 0 &0 \\ \hline
     \bfseries       4 & 0 & 0 & 0 &	0 & 50 &0 & 0 &0 & 0 &0 \\ \hline
     \bfseries       5 & 0 & 0 & 0 &	0 & 0 & 50 & 0 &0 & 0 &0 \\ \hline
      \bfseries      6 & 0 & 2 & 0 &	0 & 0 &0 & 48 &0 & 0 &0 \\ \hline
     \bfseries       7 & 0 & 0 & 0 &	0 & 0 &0 & 0 &50 & 0 &0 \\ \hline 
      \bfseries      8 & 0 & 0 & 0 &	0 & 0 &0 & 0 &0 & 50 &0 \\ \hline
      \bfseries      9 & 0 & 0 & 0 &	0 & 0 &0 & 0 &0 & 0 &50 \\ \hline
		\end{tabular}}{\label{confusion matrix}}
	\end{small}
\end{table}


\section{Conclusion}
The OCR is a benchmark problem in pattern recognition and has wide commercial interest. The work is based on the dataset CMATERDB 3.3.1. Das et al. performed the first work done on this dataset while using MLP in recognition. This model gains an accuracy of 93.8\%. The work in \cite{ashiquzzaman} is based on the work of \cite{Nibaran} and uses CNN as its model. This model uses dropout regularization and ReLU as activation function, and for the final output softmax function is used in the output layer. This model achieves an accuracy of 97.3\%. This paper proposes a modification to the model proposed in \cite{ashiquzzaman}. We add data augmentation to prevent overfitting, as well as change the activation function from ReLU to ELU to prevent vanishing gradient problem and make the learning more uniform. After adopting all these changes, the proposed model achieves an accuracy of 99.4\% which is better than any of the previous works on this dataset.

\section*{Acknowledgment}
The authors would like to thank department of Computer Science and Engineering, University of Asia Pacific for providing support in this research.

\bibliography{bib}{}
\bibliographystyle{IEEEtran}
\end{document}